\documentclass[runningheads]{llncs}

\usepackage{graphicx}
\usepackage{graphics}
\usepackage{wrapfig}
\usepackage{float}
\usepackage{textcomp}

\usepackage{graphicx}
\begin{document}
\title{Improved Padding in CNNs for Quantitative Susceptibility Mapping}
\titlerunning{Improved Padding in CNNs for QSM}
%


\author{Juan Liu}

\institute{{juan.liu313@gmail.com}}

\maketitle              
\begin{abstract} 
Recently, deep learning methods have been proposed for quantitative susceptibility mapping (QSM) data processing - background field removal, field-to-source inversion, and single-step QSM reconstruction. However, the conventional padding mechanism used in convolutional neural networks (CNNs) can introduce spatial artifacts, especially in QSM background field removal and single-step QSM which requires inference from total fields with extreme large values at the edge boundaries of volume of interest. To address this issue, we propose an improved padding technique which utilizes the neighboring valid voxels to estimate the invalid voxels of feature maps at volume boundaries in the neural networks. Studies using simulated and in-vivo data show that the proposed padding greatly improves estimation accuracy and reduces artifacts in the results in the tasks of background field removal, field-to-source inversion, and single-step QSM reconstruction. 

\keywords{Quantitative susceptibility mapping  \and Padding \and Deep learning.}
\end{abstract}

\section{Introduction}
In quantitative susceptibility mapping (QSM), tissue susceptibility is quantitatively estimated by extracting Larmor frequency offsets from complex MR signals to solve for the source tissue susceptibility \cite{wang2015quantitative}. QSM processing usually involves a series of post-processing procedures, including (1) estimating the magnetic field from the raw MR phase data, (2) eliminating the background field contributions from outside a region of interest (ROI), e.g. brain, such as air-tissue interface, to determine the local tissue field, (3) solving the field-to-source inverse problem to get the tissue susceptibility distribution. In single-step QSM, the tissue susceptibility is directly estimated from the total field without the background field removal. 

\begin{figure}[H]
\begin{center}
\includegraphics[width=.9\textwidth]{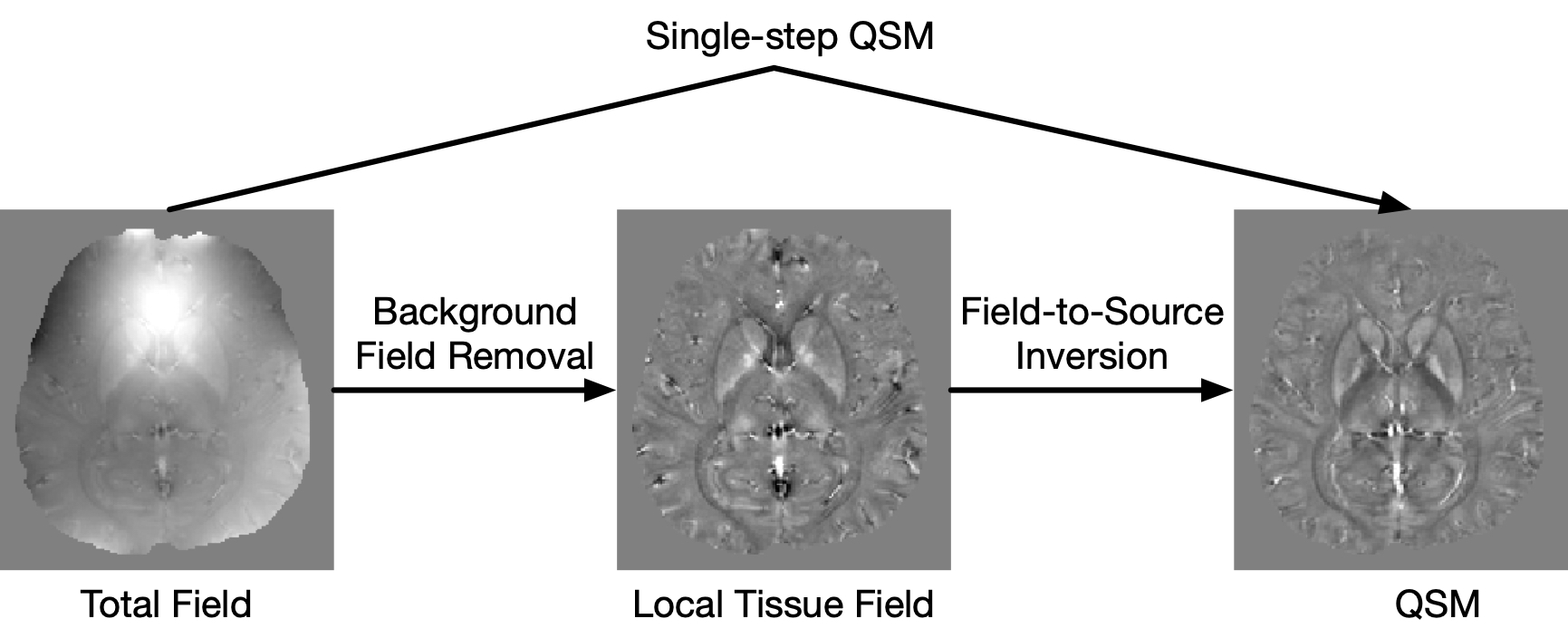}
\caption{Illustration of QSM processing.}
\label{fig:QSMprocessing}
\vspace{-20pt}
\end{center}
\end{figure}

Both background field removal and field-to-source inversion require to solve ill-posed inverse problems. Conventional methods for background field removal have the difficulties of accurately estimating the local tissue field. For field-to-source inversion, the noise amplification often causes large susceptibility quantification errors that appear as streaking artifacts in the reconstructed QSM, especially in massive hemorrhagic regions. Fig.\ref{fig:QSMprocessing} illustrates the QSM data processing. In the total field, large susceptibility variations exist in skull and paranasal sinuses introduce strong background field, especially close to the brain boundaries.

With the development of deep learning (DL), recent efforts have demonstrated the advantages of DL for QSM in background field removal\cite{bollmann2019sharqnet,liu2019deep}, field-to-source inversion\cite{yoon2018quantitative,bollmann2019deepqsm,jung2020exploring,gao2021xqsm,chen2019qsmgan}, and single-step QSM\cite{wei2019learning}. All these methods utilized U-Net \cite{ronneberger2015u} like architecture with convolutional layers, max-pooling layers, and deconvolutional layers etc. In these networks, the shape of the neural network outputs keeps the same shape of the neural network inputs. Therefore, padding is always used to overcome the shrinking outputs and losing information on corners of the feature maps. Conventional padding techniques includes zero-padding, symmetric padding, and reflective padding. In \cite{chen2019qsmgan}, the cropped outputs were used to calculate the loss since the effective receptive field of voxels near the patch edge is smaller than that of voxels near the patch center which causes the inaccurate estimation of voxels near the patch edge. However, in the prediction process, the voxels near volume boundaries still have missing information for accurate quantification. 

Recent studies \cite{alsallakh2020mind,islam2020much,kayhan2020translation} have found that the padding mechanism can encode spatial location and introduce spatial artifacts in CNNs. In QSM, the invalid voxels outside of ROIs could introduce inaccurate learning close to volume boundaries and introduce spatial artifacts in the final results. Through investigation on conventional padding techniques, we found that these conventional padding techniques do not work well in CNNs for QSM. Especially in CNNs for background field removal and single-step QSM, the strong background field at the volume boundaries could introduce severe artifacts in the results. 

To address this problem, a new padding mechanism was proposed. The padding mechanism uses the neighboring voxels of feature maps to estimate the invalid voxels at the boundaries of feature maps. We used simulated and in-invo data for quantitative evaluation on the tasks of background field removal, field-to-source inversion, and single-step QSM tasks. 

\section{Method}

Let $X$ are the feature values (voxels values) and $M$ is the corresponding binary mask. First, a convolution with all-one 3x3x3 kernel was the padded binary mask to get the scaling factor $1/sum(M)$ which applies appropriate scaling to adjust for the varying amount of invalid inputs. Second, a convolution with all-one 3x3x3 kernel was the each feature map to get the average value of valid neighboring voxels for the invalid voxels. For better generalization, the convolution kernels for feature maps and binary mask were set trainable, which were initialized with all-one. After each convolution operation, the mask was not updated.

\begin{figure}[H]
\vspace{-5pt}
\begin{center}
\includegraphics[width=\textwidth]{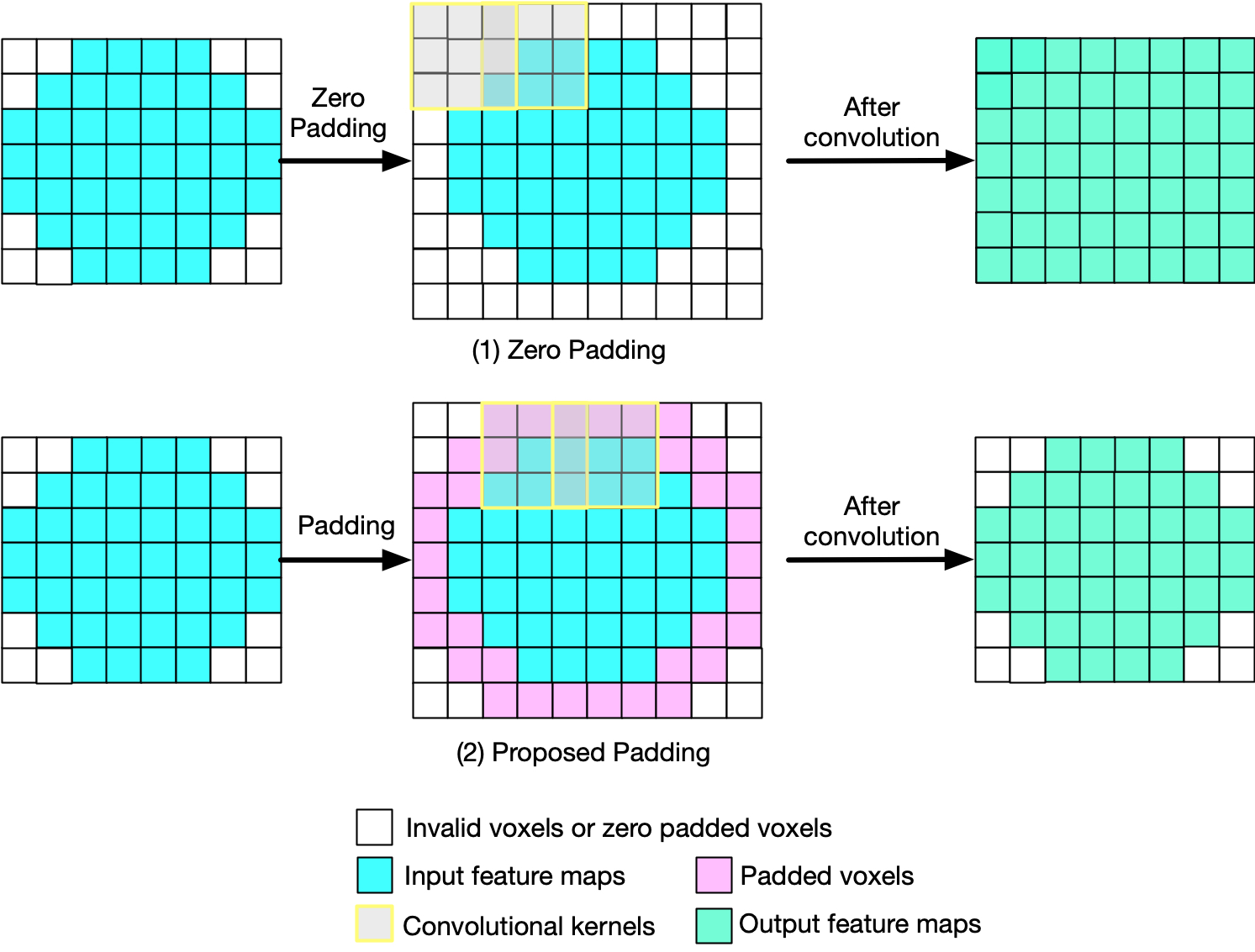}
\caption{Illustration of the zero-padding and the proposed padding. In the input feature maps, the valid voxels show blue and invalid with white. In the proposed padding, the invalid voxels at the boundaries are estimated from its neighboring valid voxels during network training before convolution, with color pink. After convolution, the feature maps at valid positions are updated.}
\label{fig:pad_Howto}
\vspace{-20pt}
\end{center}
\end{figure}

\section{Experiments}
\textbf{Synthetic Data} We used the COSMOS result of 2016 QSM reconstruction challenge to generate the simulated data. We applied random elastic transform, contrast change, and adding pseudo high susceptibility sources to augment the single QSM. The background field were simulated by placing random background susceptibility sources with large susceptibility values outside the brain. The dipole convolution were then performed to get the induced total field and local field from the susceptibility distribution. 

100 datasets with matrix size 160x160x160 and voxel size 1.0x1.0x1.0mm$^3$ were generated for network training tasks for background field removal, field-to-source inversion, and single-step QSM. 

The network adopted a 3D U-Net like architecture, using patch-based training with patch size 96x96x96, and L2 loss. We compared four padding mechanisms - (1) zero padding, (2) reflective padding, (3) symmetric padding, and (4) the proposed one. 

100 testing datasets were generated using the same way as training data. The prediction results were evaluated with respect to the ground truth using quantitative metrics - peak signal-to-noise ratio (PSNR), normalized root mean squared error (NRMSE), high frequency error norm (HFEN), and structure similarity (SSIM) index.    

\textbf{In-vivo Data} 9 QSM datasets were acquired using 5 head orientations and a 3D single-echo GRE scan with isotropic voxel size 1.0x1.0x1.0 mm$^3$ on 3T MRI scanners. QSM data processing was implemented as following, offline GRAPPA \cite{griswold2002generalized} reconstruction to get magnitude and phase images from saved k-space data, coil combination using sensitivities estimated with ESPIRiT \cite{uecker2014espirit}, BET (FSL, Oxford, UK) \cite{smith2002fast} for brain extraction, Laplacian method \cite{li2011quantitative} for phase unwrapping, and RESHARP \cite{wu2012whole} with spherical mean radius 4mm for background field removal. COSMOS results were calculated using the 5 head orientation data which were registered by FLIRT (FSL, Oxford, UK)\cite{jenkinson2002improved,jenkinson2001global}.  

In network training for background field removal, the RESHARP results were used as the training label. Since RESHARP local fields have brain erosion, the input total field used the eroded volume. Leave-one-out cross validation was used. For each dataset, total 40 scans (8*5) from other 8 datasets were used for training. The network was trained on patch-based with patch size 96x96x96, L2 loss. For field-to-source inversion, the COSMOS maps were used as training labels with network input of local tissue field. Leave-one-out cross validation was used, only using the normal head position scans for training. For single-step QSM reconstruction, the COSMOS maps were used as the training label with the total field as the network input.

\section{Results}

\begin{figure}[H]
\vspace{-0pt}
\begin{center}
\includegraphics[width=\textwidth]{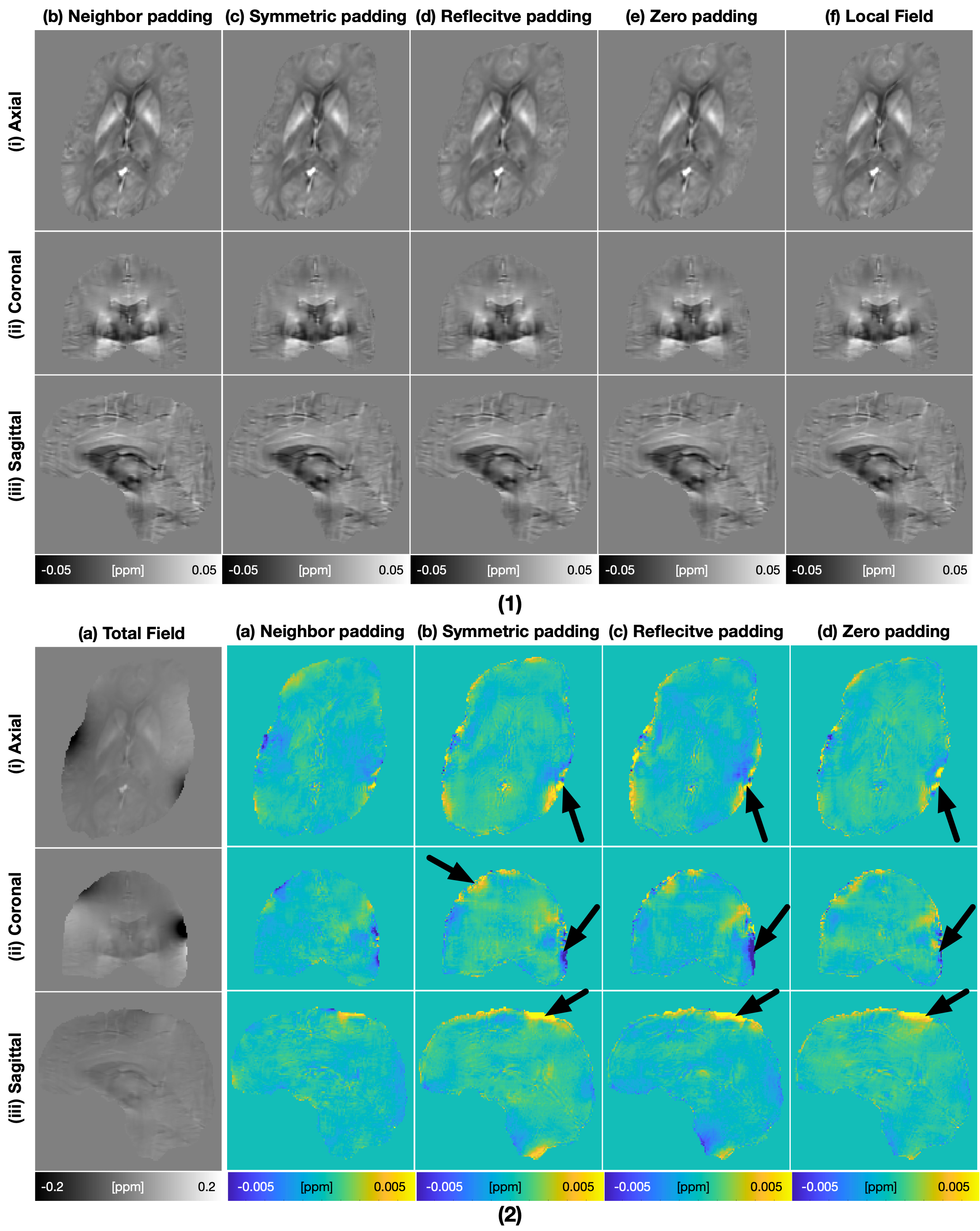}
\caption{Comparison of background field removal performance on a representative synthetic testing data. From the residual error map (2), the proposed padding method has less residual errors, especially close to brain boundaries.}
\label{fig:pad_background}
\vspace{-20pt}
\end{center}
\end{figure}

\textbf{Synthetic Data}
Table \ref{tab:uQSM1}  displays the quantitative evaluation results. In the all three tasks, the proposed method achieved the best scores in all metrics. Fig \ref{fig:pad_background} displays the background field removal results from networks with different padding techniques. From the residual maps, the proposed padding technique have obvious less residual errors, especially close to brain boundaries with strong background field.  

\begin{table}[H]
\vspace{-5pt}
\centering
\caption{\label{tab:uQSM1} Means and standard deviations of quantitative performance metrics on 100 synthetic testing data.}
  {\begin{tabular}{lllll}
  {} &  \bfseries{PSNR (dB)} &  \bfseries{NRMSE ($\%$)} & \bfseries{HFEN ($\%$)} & \bfseries{SSIM (0-1)}\\

 \bfseries{Background field removal}\\
 {zero padding} &  {50.3$\pm$5.3} &  {12.8$\pm$2.3} &  {11.5$\pm$2.6} &  {0.998$\pm$0.001}\\
 {symmetric padding} &  {49.1$\pm$5.4} &  {14.7$\pm$2.3} &  {12.5$\pm$2.4} &  {0.998$\pm$0.002}\\
 {reflective padding} &  {{49.1$\pm$5.3}} &  {{14.7$\pm$2.4}} &  {{12.7$\pm$2.7}} &  {$0.998\pm0.002$}\\
 {neighbor padding} &  {\textbf{52.6$\pm$5.2}} &  {\textbf{9.9$\pm$1.6}} & {\textbf{8.9$\pm$1.8}} &
 {\textbf{0.999$\pm$0.001}}\\
 
 \bfseries{Field-to-source Inversion}\\
 
  {zero padding} &  {45.3$\pm$4.2} &  {19.0$\pm$1.8} &  {19.2$\pm$1.5} &  {0.984$\pm$0.010}\\
 {symmetric padding} &  {45.0$\pm$4.3} &  {19.6$\pm$1.8} &  {19.9$\pm$1.6} &  {0.984$\pm$0.010}\\
 {reflective padding} &  {{44.7$\pm$4.3}} &  {{20.3$\pm$1.8}} &  {{20.8$\pm$1.7}} &  {$0.983\pm0.010$}\\
 {neighbor padding} &  {\textbf{46.0$\pm$4.2}} &  {\textbf{17.4.2$\pm$1.7}} &  {\textbf{17.3$\pm$1.2}} &  {\textbf{0.986$\pm$0.009}}\\
 
 \bfseries{Single-step QSM}\\
  {zero padding} &  {42.6$\pm$4.4} &  {25.8$\pm$2.2} &  {27.7$\pm$2.7} &  {0.974$\pm$0.016}\\
 {symmetric padding} &  {42.3$\pm$0.6} &  {26.8$\pm$2.8} &  {28.9$\pm$3.0} &  {0.973$\pm$0.017}\\
 {reflective padding} &  {{42.1$\pm$4.5}} &  {{27.4$\pm$2.4}} &  {{29.2$\pm$2.9}} &  {$0.972\pm0.017$}\\
 {neighbor padding} &  {\textbf{44.6$\pm$4.3}} &  {\textbf{20.6$\pm$1.9}} &  {\textbf{21.4$\pm$1.9}} &  {\textbf{0.983$\pm$0.010}}\\
  \end{tabular}}
  \vspace{-15pt}
\end{table}

\begin{figure}[H]
\vspace{-5pt}
\begin{center}
\includegraphics[width=\textwidth]{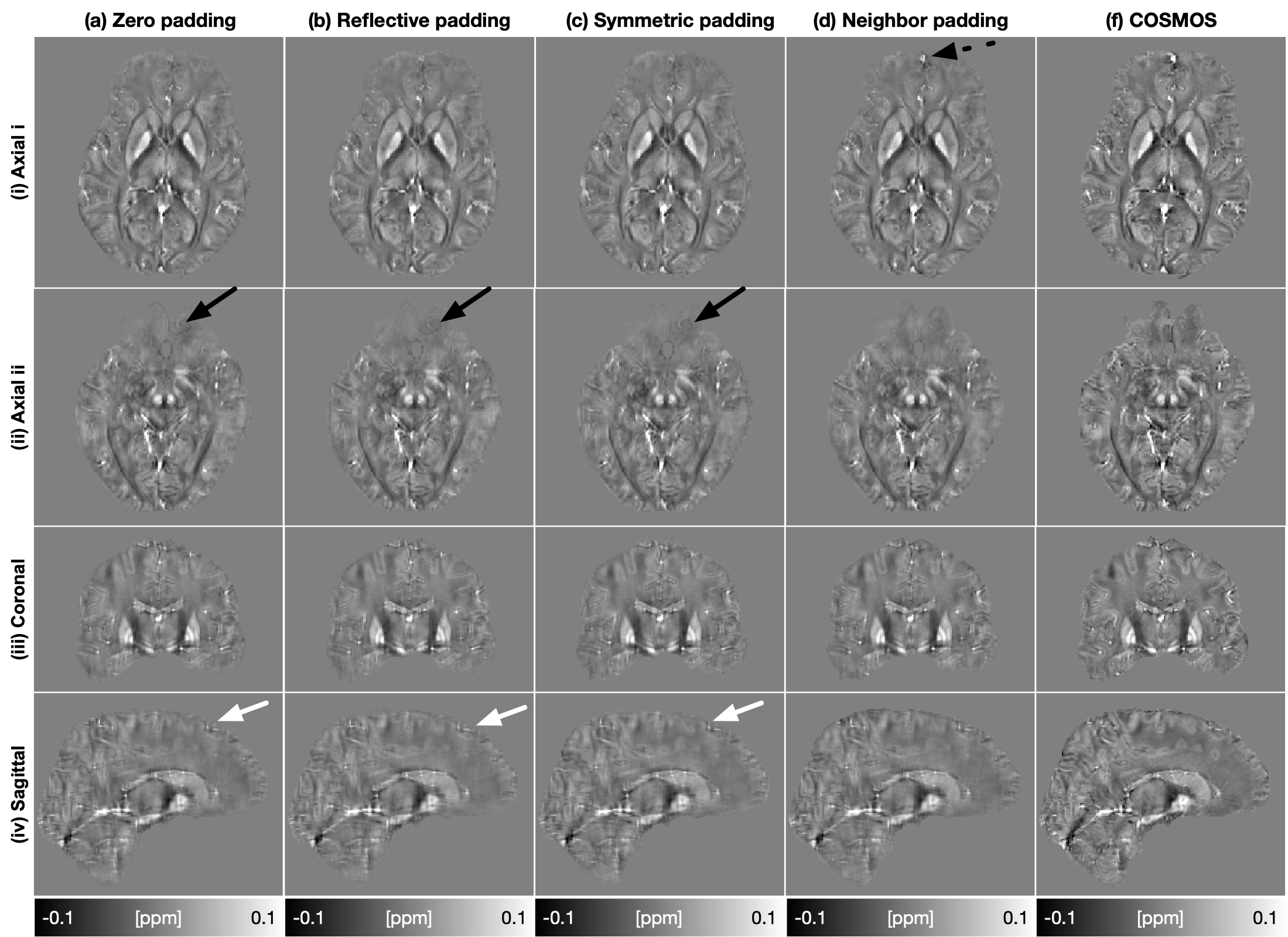}
\caption{Comparison of single-step QSM performance on an in-vivo data. In the results of zero padding, reflective padding and symmetric padding show the obvious artifacts (ii, black arrows) and jagged-like artifacts at the boundaries (iv, white arrows).}
\label{fig:pad_ssqsm_invivo}
\vspace{-20pt}
\end{center}
\end{figure}

\textbf{In-vivo Data} Table \ref{tab:9QSM}  displays the quantitative evaluation results. In the tasks of background field removal, field-to-source inversion, and single-step QSM, the proposed method achieved the best scores in all metrics. Fig \ref{fig:pad_ssqsm_invivo} display the background field removal results from networks with different padding techniques. From the residual maps, the result of the proposed padding technique have obvious residual errors especially close to brain boundaries. 

\begin{table}[H]
\centering
\caption{\label{tab:9QSM} Means and standard deviations of quantitative performance metrics of cross-validation on multi-orientation data.}
  {\begin{tabular}{lllll}
  {} &  \bfseries{PSNR (dB)} &  \bfseries{NRMSE ($\%$)} & \bfseries{HFEN ($\%$)} & \bfseries{SSIM (0-1)}\\
 
 \bfseries{Background field removal}\\
 {zero padding} &  {35.0$\pm$0.9} &  {26.4$\pm$2.2} &  {24.2$\pm$1.9} &  {0.992$\pm$0.002}\\
 {symmetric padding} &  {35.2$\pm$0.8} &  {27.2$\pm$2.0} &  {24.7$\pm$1.9} &  {0.991$\pm$0.002}\\
 {reflective padding} &  {{35.0$\pm$0.9}} &  {{27.4$\pm$1.9}} &  {{25.0$\pm$1.9}} &  {0.992$\pm$0.002}\\
 {neighbor padding} &  {\textbf{37.8$\pm$0.8}} &  {\textbf{19.7$\pm$1.0}} & {\textbf{18.4$\pm$1.1}} &
 {\textbf{0.995$\pm$0.001}}\\
 
 \bfseries{Field-to-source Inversion}\\
  {zero padding} &  {{48.76$\pm$0.69}} &  {{49.5$\pm$3.3}} &  {{42.6$\pm$3.0}} &  {{0.912$\pm$0.0140}}\\
 {symmetric padding} &  {48.78$\pm$ 0.68} &  {49.4$\pm$3.3} &  {42.3$\pm$2.9} &  {0.912$\pm$ 0.013}\\
 {reflective padding} &  {{48.76$\pm$0.66}} &  {{49.5$\pm$3.2}} &  {{42.5$\pm$2.8}} &  {0.911$\pm$0.014}\\
 {neighbor padding} &   {\textbf{48.83$\pm$0.69}} &  {\textbf{49.1$\pm$3.3}} &  {\textbf{42.1$\pm$2.9}} &  {\textbf{0.913$\pm$0.012}}\\
 
 \bfseries{Single-step QSM}\\
  {zero padding} &  {47.7$\pm$0.8} &  {55.9$\pm$3.7} &  {49.0$\pm$3.4} &  {0.897$\pm$0.015}\\
 {symmetric padding} &  {47.7$\pm$0.7} &  {56.0$\pm$3.3} &  {49.1$\pm$3.1} &  {0.897$\pm$0.015}\\
 {reflective padding} &  {{47.6$\pm$0.8}} &  {{56.5$\pm$3.4}} &  {{49.6$\pm$3.2}} &  {$0.897\pm0.015$}\\
 {neighbor padding} &  {\textbf{48.1$\pm$0.7}} &  {\textbf{53.5$\pm$3.5}} &  {\textbf{46.2$\pm$3.1}} &  {\textbf{0.903$\pm$0.014}}\\
  \end{tabular}}
\end{table}

\section{Discussion}

In this work, an improved padding technique was proposed to decrease the spatial artifacts in QSM CNNs. For the patch edges, the invalid pixels was first approximated using the neighboring valid pixels before convolution. From quantitative evaluation and visual assessment on synthetic datasets and in-vivo datasets, the proposed padding technique achieved impressive performance. Especially in the tasks of background field and single-step QSM, the proposed methods showed substantial less errors than conventional padding techniques. While in the task of field-to-source inversion, the spatial artifacts is less compared to the tasks of background field and single-step QSM. 

In the tasks of background field and single-step QSM, a strong background field contamination exists close to irregular brain boundaries (tissue air interface). Therefore, it is of importance to take this account in the network design. This causes conventional padding techniques failure and introduce the spatial artifacts in the results. In the task of field-to-source inversion, the local field usually does not have strong variation close to brain boundaries, the spatial artifacts is not obvious when using conventional padding techniques. 

\section{Conclusion}

The proposed padding demonstrated better performance than conventional padding techniques in three deep learning tasks for QSM. In the tasks of background field and single-step QSM, the proposed methods significantly reduce the errors close to volume boundaries. We believe that the proposed padding technique could improve DL-based QSM techniques. 

\begin{figure}[H]
\vspace{-15pt}
\begin{center}
\includegraphics[width=\textwidth]{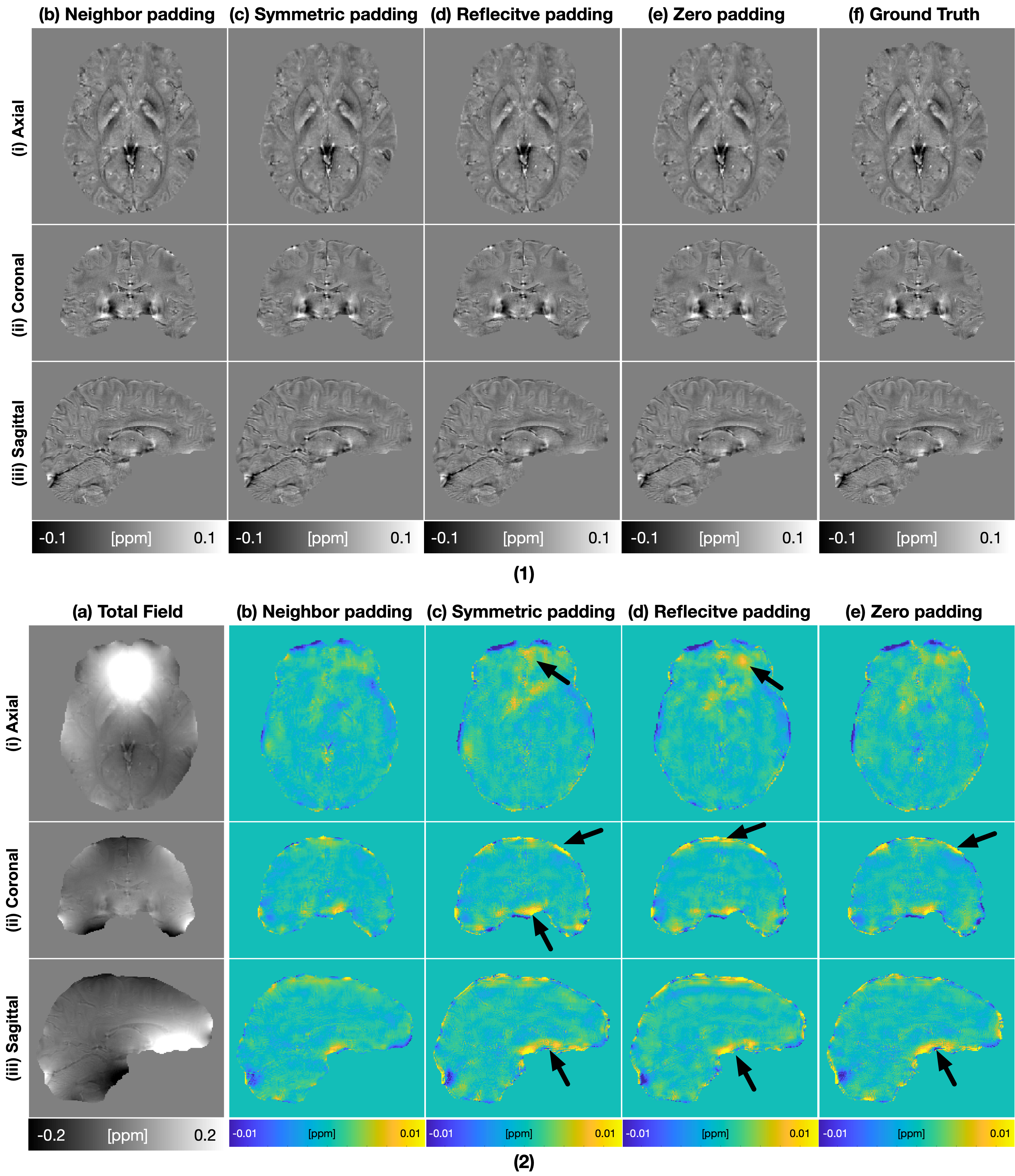}
\caption{Comparison of QSM background performance on an in-vivo data. In the results of zero padding, reflective padding and symmetric padding show the obvious artifacts (black arrows).}
\label{fig:pad_background_invivo}
\vspace{-20pt}
\end{center}
\end{figure}

\section*{Acknowledgement}
We thank Professor Jongho Lee for sharing the multi-orientation QSM datasets.

\bibliographystyle{splncs04}
\bibliography{references}






\end{document}